\documentclass{article}

    \usepackage[preprint,nonatbib]{neurips_2020}

\usepackage[utf8]{inputenc} 
\usepackage[T1]{fontenc}    
\usepackage{hyperref}       
\usepackage{url}            
\usepackage{booktabs}       
\usepackage{amsfonts}       
\usepackage{nicefrac}       
\usepackage{microtype}      
\usepackage{amsmath}
\usepackage{graphicx}
\usepackage{xcolor}

\title{Latent Video Transformer}

\author{%
  Ruslan Rakhimov\thanks{Equal contribution} \\
   Skolkovo Institute of Science and Technology \\
  \texttt{ruslan.rakhimov@skoltech.ru} \\
   \And
   Denis Volkhonskiy$^*$ \\
   Skolkovo Institute of Science and Technology \\
   \texttt{denis.volkhonskiy@skoltech.ru} \\
   \AND
   Alexey Artemov \\
   Skolkovo Institute of Science and Technology \\
   \texttt{a.artemov@skoltech.ru} \\
   \And
   Denis Zorin \\
   New York University \\
   Skolkovo Institute of Science and Technology \\
   \texttt{dzorin@cs.nyu.edu} \\
   \And
   Evgeny Burnaev \\
   Skolkovo Institute of Science and Technology \\
   \texttt{e.burnaev@skoltech.ru} \\
}

\newenvironment{itemizetight}
{\vspace{-.5\baselineskip}\begin{itemize}\setlength{\itemsep}{-.25\baselineskip}}
{\end{itemize}\vspace{-.5\baselineskip}}

\begin{document}

\maketitle

\begin{abstract}
The video generation task can be formulated as a prediction of future video frames given some past frames. Recent generative models for videos face the problem of high computational requirements. Some models require up to $512$ Tensor Processing Units for parallel training. In this work, we address this problem via modeling the dynamics in a latent space. After the transformation of frames into the latent space, our model predicts latent representation for the next frames in an autoregressive manner. We demonstrate the performance of our approach on BAIR Robot Pushing and Kinetics-600 datasets. The approach tends to reduce requirements to 8 Graphical Processing Units for training the models while maintaining comparable generation quality.
\end{abstract}
\section{Introduction}

Video prediction and generation is an important problem with a lot of down-stream applications: self-driving, anomaly detection, timelapse generation \cite{timelapse_generation}, animating landscape \cite{animating_landscape} etc. The task is to generate the most probable future frames given several initial ones.


Recent advances in generative learning allow generation of realistic objects with high quality: images, text, and speech. However, video generation is still a very challenging task. Even for short videos ($16$~frames) of low resolution, neural networks require up to $512$ Tensor Processing Units (TPUs) \cite{dvdgan2} for parallel training. Despite this, the quality of the generated video remains low.

In this work, we introduce a Latent Video Transformer\footnote{Source code is available at \url{https://github.com/rakhimovv/lvt}}. We combine the idea of representation learning and recurrent video generation. Instead of working in pixel space, we conduct the generation process in the latent space. Our model tends to significantly reduce computational requirements without significant deterioration in quality.

The key novelty in our model is the usage of a discrete latent space \cite{vqvae}. It allows us to represent each frame as a set of indices. Thanks to discrete representation 
we can use autoregressive generative models and other approaches from natural language processing.

We analyzed the results of our model on two datasets: BAIR Robot Pushing \cite{bair_dataset} and Kinetics 600 \cite{kinetics600_dataset}. On both datasets, we obtained quality comparable to state-of-the-art methods.

To summarize, our contributions are as follows:

\begin{itemizetight}
    \item We proposed a new autoregressive model for video generation, that works in the latent space rather than pixel space;
    \item We reduced computational requirements comparing to previously proposed methods.
\end{itemizetight}




\section{Related work}

\subsection{Video Generation}

Video generation is a long-standing problem. One can formulate it in different ways. Future video prediction, unconditional video synthesis or video to video translation \cite{vid2vid, video_from_semantic, vid2vid++}. There exist other rare setups like generating video from one image \cite{singan}.
Video prediction and unconditional video synthesis have been addressed for a long time and the solutions include recurrent models \cite{ulpi, predrnn++, vpb, favp}, VAE-based models \cite{sv2p, uldr, savp, ldrp, svg}, autoregressive models \cite{vm, ulstm, clstm, video_pixel_networks, vt, axial_transformer}, normalizing flows \cite{kumar}, GANs \cite{dmsv, vgan, tgan, mocogan, pvgan, tganv2, dvdgan, dvdgan2} and optical flow \cite{ptrucean, katsunori}.

In the first attempts, fully deterministic models have been used. Later generative models were applied. Similar to image generation, video generative models inherited similar benefits and drawbacks. Variational autoencoder (VAE)-based models try to model videos in latent space but produce blurry results. Later GANs were applied to address those issues, but they suffer from mode-dropping behavior. Some models \cite{savp} try to combine VAE and GANs.

The recent state-of-the-art approaches DVD-GAN-FP \cite{dvdgan} and its modification TRIVD-GAN-FP \cite{dvdgan2} follow the success of BigGAN \cite{biggan}. They use 2D residuals blocks for independent frames prediction with Convolutional Gated Recurrent units between frames.

Another branch of generative models is autoregressive (AR) models. PixelCNN \cite{pixelcnn, pixelcnn++} generates new images by producing a new pixel value conditioning on previous (seen, already generated) ones in the raster-scan order. Later, PixelSnail \cite{pixelsnail} increased the quality of generated samples by utilizing an attention mechanism. Recently, such an approach was applied to video generation \cite{video_pixel_networks, vt}. The latest work, Video Transformer \cite{vt}, utilizes autoregressive video generation along with subscaling \cite{pixelcnn_subscale} and attention mechanism \cite{transformer}.

The main challenge of AR models is a generation speed. Even though the latest AR model (VideoTransformer \cite{vt}) applied the subscaling mechanism \cite{pixelcnn_subscale}, introduced block-local attention, the generation speed is still quite slow.

Also, DVD-GAN-FP, TRIVD-GAN-FP, Video Transformer (VT) --- they are all suffering from significant resource requirements, even for generating low-resolution video with frames of size 64x64. For instance, VT needs 128 TPUs and 1M steps for training.

Our work is in the field of autoregressive models and follows the setup of VideoTransformer \cite{vt}. The key novelty is that we mitigate GPU memory consumption and accelerate inference speed by working in a discrete latent space.

\subsection{Discrete latent space}

Autoencoder is a neural network trained in a self-supervised manner. Autoencoder takes an input (image, text, audio, etc.) and transfers (encodes) it into a more compact latent representation. The learning consists of finding such an encoder and decoder so that we can encode and decode the input as closely as possible.

Usually, latent space is continuous. However, some works like VQ-VAE \cite{vqvae}, VQ-VAE2 \cite{vqvae2} model it as discrete with a categorical distribution inside. They demonstrated good reconstruction and generation quality. As generating from uniform distribution directly produced inferior results, autoregressive models were applied to learn the prior inside the latent space.

We follow this pipeline but for video modeling. First, encoding conditioning frames to discrete latent space, generate new (latent) frames using an autoregressive model, and decode the generated frames back to pixel space. Parallel to this work, a similar pipeline was applied to audio generation (Jukebox \cite{jukebox}).

Discrete latent space also occurred to be useful in other works. Discrete quantization was added to a discriminator in GAN \cite{quantization_gan}. \cite{lt} uses discrete variables to increase the speed of the autoregressive model for neural machine translation.
\section{Latent Video Transformer}

Consider a video $X$ to be a sequence of $T$ frames $\{x_t\}_{t=1}^{T}$. Each frame $x_t \in \mathbb{R}^{H \times W \times 3}$ has height $H$, width $W$ and $3$ RGB channels. Given the first $T_0$ frames, the goal is to generate the remaining $T-T_0$ frames. For this purpose, we propose a model: Latent Video Transformer (LVT). In general, it consists of two parts: a frame autoencoder and an autoregressive generative model.

We use the frame autoencoder to learn a compact latent representation so that we can transfer the task of video modeling from the pixel space to the latent space. The recurrent model is then used for generating new frames. The key novelty compared to existing models that operate in the latent space is a discrete structure of the latent space. 
Discrete representation helps us to use autoregressive generative models and other approaches, tailored for working for discrete data, e.g. those used for natural language processing.

\subsection{Frame Autoencoder}

We train a frame autoencoder to transfer individual images (frames) to latent space. The particular choice of the autoencoder is VQ-VAE \cite{vqvae} --- variational autoencoder with discrete latent space.

\begin{figure}[h!]
\centering
\includegraphics[width=0.7\textwidth]{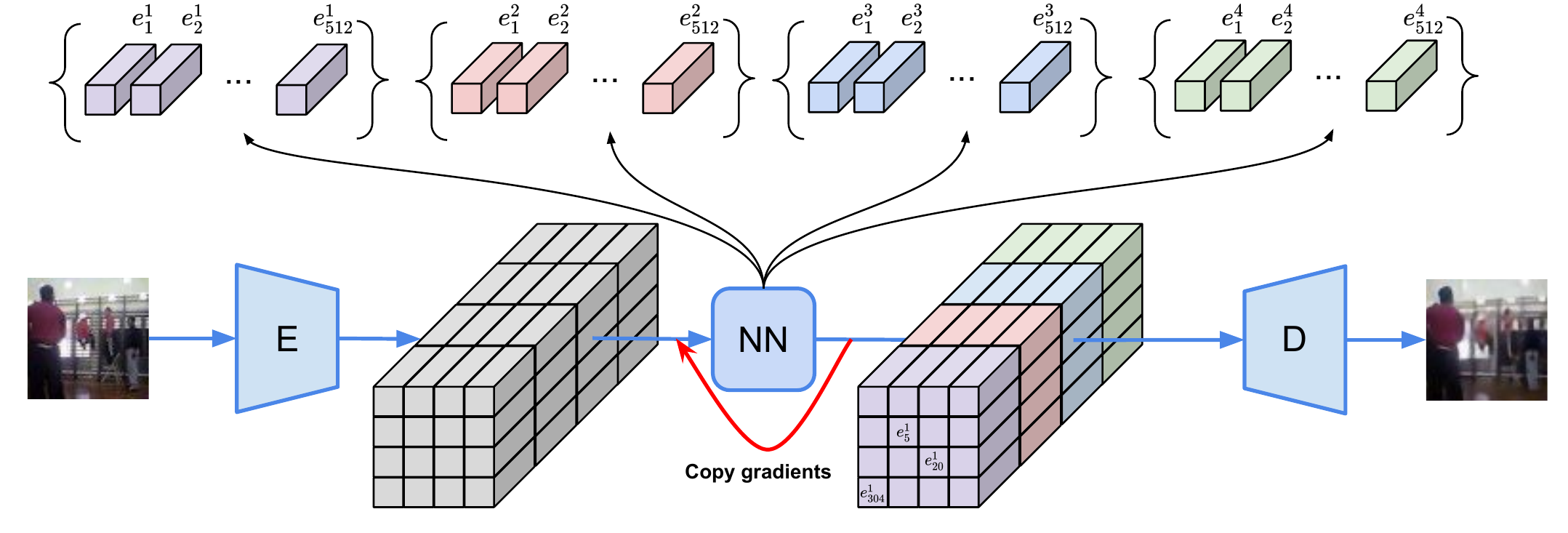}
\caption{Frame autoencoder architecture. An input image is passed through the encoder and split along the channel dimension into $n_c=4$ parts. Then we map pixels in each part to the corresponding nearest embeddings in the codebook. These nearest embeddings are then passed as an input to the decoder.}
\label{fig:ae}
\end{figure}

VQ-VAE (see Fig. \ref{fig:ae}) learns to encode an input image $x \in \mathbb{R}^{H \times W \times 3}$ using a codebook $e\in\mathbb{R}^{K\times D}$, where $K$ denotes the codebook size (i.e., latent space is K-way categorical) and $D$ represents the size of an embedding in the codebook.

In general VQ-VAE consists of an \textit{encoder} which encodes the image into more compact representation $E(x)=z_e(x)\in\mathbb{R}^{h \times w \times D}$;  \textit{a bottleneck}, that discretizes each pixel by mapping it to its nearest embedding $e_i$ from the codebook and produces $z(x)\in[K]^{h \times w \times 1}$; \textit{a decoder} $D$ takes as input discrete latent codes $z(x)$, maps indexes to corresponding embeddings, and decodes the result of mapping $z_q(x)\in\mathbb{R}^{h \times w \times D}$ back to input pixel space.

VQ-VAE is trained with the following objective:

\begin{equation}
\label{eq:ae_loss}
    L =\| x - D(z_q(x))\|^2 + \| z_e(x) - \text{sg}[e]\|^2,
\end{equation}

where sg[] is the stop gradient operator, which returns its argument during the forward pass and zero gradients during backward pass. The first term is a reconstruction loss, and the second term is regularization term to make the encodings less volatile. We use EMA updates over the codebook variables.

\textbf{Decomposed Vector Quantization.} If the size $K$ of the codebook $e$ is large, then the model tends to index collapse. It means that some embedding vector $e_i$ is close to a lot of encoders outputs. In this case, it receives a strong update signal \cite{lt}. As a result, the model would use only a limited number of vectors from $e$.

In order to overcome this issue, we exploit Sliced Vector Quantization \cite{lt}. We introduce several codebooks $\{e^j\in\mathbb{R}^{K\times D/n_c}\}_{j=1}^{n_c}$ and split the output of encoder $z_e(x)$ along the channel dimension into $n_c$ parts with individual codebook per each part (see Figure \ref{fig:ae}). The output from discretization bottleneck in this case is $z\in[K]^{h \times w \times n_c}$.

\subsection{Latent Video Generator}

Frame encoder transforms the first $T_0$ frames to a discrete representation $Z_0\in[K]^{T_0\times h \times w \times n_c }$.

The autoregressive model is used to generate new $T-T_0$ frames conditioned on $Z_0$. As such model, we use the Video Transformer \cite{vt}, autoregressive video generative model, but apply it in the latent space in contrast to the pixel space in the original paper. Next, we describe the architecture of a video transformer. We abuse notation and refer to a latent representation of a video also as \textit{video} and individual elements of it as \textit{frames} and \textit{pixels}. For exhaustive architecture details, we refer the reader to the original paper \cite{vt}.

\begin{figure}[h!]
\centering
\includegraphics[width=0.6\textwidth]{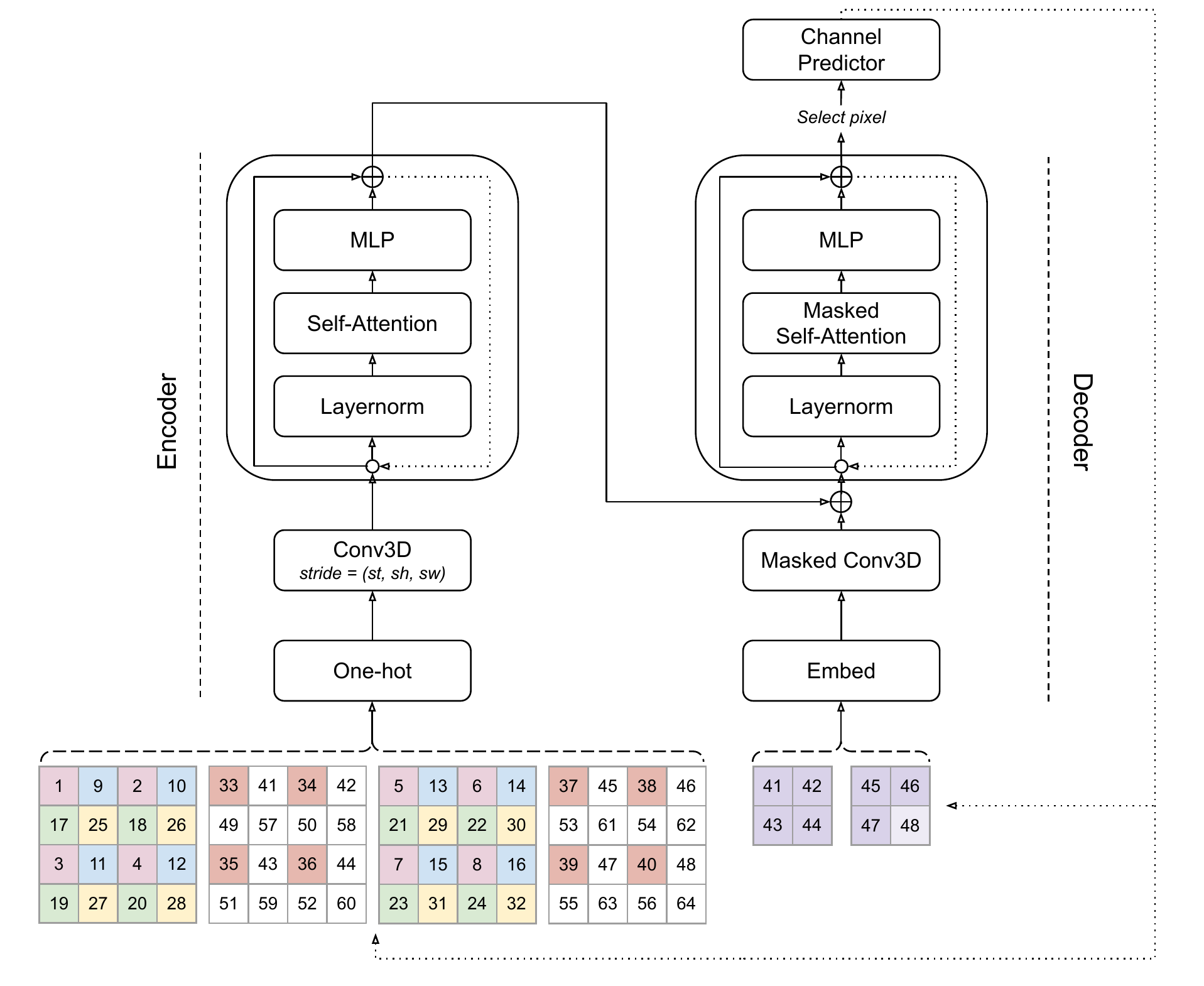}
\caption{Video Transformer adapted to latent codes. Numbers represent generation order. Pixels are colored if they are already generated. White-colored pixels are zero-padded. Pixels with the same color belong to the same slice. The example represents the generation of the last pixel of slice $Z_{(1,0,1)}$ for a video of size $(t,h,w)=(4,4,4)$ and $(s_t,s_h,s_w)=(2,2,2)$.}
\label{fig:lvt}
\end{figure}


The model takes as input a tensor $Z\in[K]^{T \times h \times w \times n_c}$ and primes the generation process on first $T_0$ given frames, i.e. $Z_{:T_0,:,:,:}=Z_0$. The other frames could be randomly filled as the generation process is conditioned only on already generated or priming pixels.

First, the model utilizes the idea of subscaling \cite{pixelcnn_subscale}: let's generate a video as a sequence of \textit{non-overlapping} slices. After defining a subscale factor $\textbf{s} = (s_t, s_h, s_w)$, it divides video into $s = s_t s_h s_w$ slices of size $T/s_t \times h/s_h \times w/s_w$. The generation process happens slice by slice, pixel by pixel inside one slice, channel by channel for one pixel:

$$p(Z)=\prod_{i=0}^{Thw-1} \prod_{k=0}^{n_{c}-1} p\left(Z_{\pi(i)}^{k} | Z_{\pi(<i)}, Z_{\pi(i)}^{<k}\right)$$

Pixels in each slice $Z_{(a, b, c)}$ are generated in raster-scan order and slices are generated in the subscale order: $Z_{(0, 0, 0)}, Z_{(0, 0, 1)}, \ldots, Z_{(s_{t} - 1, s_{h} - 1, s_{w} - 1)}$.

The model follows the original Transformer \cite{transformer} and consists of an encoder and a decoder. To generate a new pixel value inside slice $Z_{(a, b, c)}$, firstly, the encoder outputs the representation of already generated slices $Z_{<(a, b, c)}$. This representation goes to the decoder, which mixes it with a representation of already generated pixels inside a current slice $Z_{(a, b, c)}$. This autoregressive order is preserved by padding input video inside the encoder, and masking used in convolutions and attention inside the decoder. After generating a new pixel value, we replace the respective padding with the generated output and repeat the generation process recursively. The generation process in case of spatiotemporal ($s_t>0, s_h>0, s_w>0$) subscaling can be seen at Fig. \ref{fig:lvt}.

Finally, when the generation process is done, the frame decoder takes as input $Z\in[K]^{T \times h \times w\times n_c}$ (now all values are valid), maps it to already learned embeddings $Z_q\in\mathbb{R}^{T \times h \times w\times D}$ and decodes it back frame by frame to an original pixel space $X\in\mathbb{R}^{T \times H \times W \times 3}$.




\section{Experiments}

\subsection{Experimental setup}

We model the videos of length $T=16$ and spatial size $64\times64$ similar to the setup of prior works in this field \cite{dvdgan, vt}.

\textbf{Measures of quality.} Video prediction is a challenging problem, as there are many possible future outcomes for given conditioning frames. Therefore conventional metrics as Peak Signal-to-Noise Ratio (PSNR) and Structural Similarity Index Measure (SSIM) that require ground truth correspondence were later displaced by the better-suited metric --- Fréchet Video Distance (FVD) \cite{fid}. FVD applies idea from Fréchet Inception Distance \cite{fid} for videos and computes Fréchet distance between real and generated samples based on statistics calculated on logits from action-recognition Inception3D network trained on Kinetics-400 \cite{kinetics400_dataset} dataset. The metric was shown to better correlate with human perception, than previously used ones. 

We also report bits per dimension\footnote{Due to the nature of frame autoencoder, it is not possible to compute bits/dim directly. We provide bits/dim in the latent space. Other works provide this metric for the pixel space. One can consider bits/dim in the latent space as the lower bound to real bits/dim on images.} (bits/dim) --- negative $\log_2$-probability averaged across all generated (latent) pixels and channels.


\textbf{Frame autoencoder.} The encoder contains two strided convolutional layers with ReLU activation function, stride $2$ and kernel size $4\times4$, followed by a convolution layer with kernel size $3\times 3$, the same padding, followed by two residual blocks (ReLU, $3\times3$ conv, ReLU, $1\times1$ conv). The decoder has a symmetrical structure containing two residual blocks, followed by two transposed convolutions with stride $2$ and window size $4\times4$. For Kinetics-600 dataset, we use four residual blocks instead of two both in encoder and decoder.

In our experiments we explore two setups for the codebook structure: (the default one) $n_c=1$, $K=512$ and $n_c=4$, $K=2048$. The embedding dimension inside codebook is $D=256$. The encoder and discretization bottleneck converts $64\times64\times3$ RGB image into $16\times16\times n_c$ discrete latent codes indices.

We trained VQ-VAE using Adam optimizer with learning rate $0.0003$ for 500K steps in case of BAIR Robot pushing dataset and 1M steps in case of Kinetics-600 dataset using a batch of 32 images.

\textbf{Latent Video Generator.} As no public code was available, Video Transformer implementation was written from scratch using the setup of a medium size model and following the implementation and training details from the original paper \cite{vt}.

Different from the original transformer, an attention block in the encoder and the decoder is not block-local and spans across the whole input. It is possible due to the reduction of the input size via VQ-VAE. In almost all our experiments we also compare different subscaling types: (i) spatiotemporal ($\mathbf{s}=(4,2,2)$), (ii) spatial ($\mathbf{s}=(1,2,2)$), and (iii) single frame ($\mathbf{s}=(T,1,1)$).

Each model was trained on $8$ Nvidia V100 GPUs for two days for 600K steps. Sampling one video of $16$ frames with $5$ priming frames takes about 30 seconds on 1 Nvidia V100 GPU, compared to one minute used by the original Video Transformer \cite{vt}. The approximate size of a latent video generator is 50M parameters.

We provide quantitative and qualitative results on two datasets: BAIR Robot Pushing \cite{bair_dataset} and Kinetics 600 \cite{kinetics600_dataset}.

\subsection{BAIR Robot Pushing}

BAIR Robot Pushing \cite{bair_dataset} dataset consists of 40K training and 256 test videos of robotic arm motion with a fixed camera position. First, we evaluate the VQ-VAE's reconstruction error with varying number of codebooks used inside the discretization bottleneck. We provide mean squared error (MSE) and FVD for reconstructed videos of 16 frames length (see Table \ref{table:BAIRvqvae}).

\begin{table}[!hbtp]
\centering
\caption{VQ-VAE performance on BAIR Robot Pushing dataset.}
\begin{tabular}{lccc}
\hline
$\mathbf{n}_c$                  & \textbf{MSE}($\downarrow$) & \textbf{FVD}($\downarrow$)     \\ \hline
$1$ & $0.0016$ & $222.71$ \\
$4$ & $0.0004$ & $47.41$ \\ \hline
\end{tabular}
\label{table:BAIRvqvae}
\end{table}

In terms of video prediction, following the setup of previous approaches \cite{vt, dvdgan, dvdgan2}, we train video generator conditioning on one frame and report metrics for videos of 16 frames. FVD and bits/dim are computed on videos with five priming frames and one priming frame accordingly (see Table \ref{table:BAIRvt}). We report the mean and standard deviations of $10$ runs.

\begin{table}[!hbtp]
\centering
\caption{Video prediction performance on BAIR Robot Pushing dataset. Best results in bold.}
\begin{tabular}{lllcccc}
\hline
\textbf{Subscaling type} & $\mathbf{n_c}$ & \textbf{Bits/dim}($\downarrow$) & \textbf{FVD}($\downarrow$) \\ \hline
Single Frame & 1 & $1.28$ & $258.89\pm2.85$ \\
Spatial & 1 & $1.79$ & $524.43\pm9.41$ \\
Spatiotemporal & 1 & $\mathbf{1.25}$ & $275.71\pm5.41$ \\ \hline
Single Frame & 4 & $1.53$ & $\mathbf{125.76 \pm 2.90}$ \\
Spatial & 4 & $2.99$ & $920.37\pm 7.71$ \\
Spatiotemporal & 4 & $1.62$ & $145.85 \pm 1.68$ \\ \hline
\end{tabular}
\label{table:BAIRvt}
\end{table}

It can bee seen that both VQ-VAE and Video Transformer demonstrate better accuracy when using four codebooks inside the discretization bottleneck. Preliminary experiments showed that a further increase of the number of codebooks would lead to overfitting. Finally, we compare our approach to others (see Table \ref{table:BAIRbench}). We also provide the \textit{baseline} solution: what if we take the last ground truth frame and use it as a prediction for all future frames.

\begin{table}[!hbtp]
\centering
\caption{Comparison of different methods for video prediction on BAIR Robot Pushing dataset.}
\begin{tabular}{lcc}
\hline
\textbf{Method}          & \textbf{bits/dim}($\downarrow$)        & \textbf{FVD}($\downarrow$)     \\ \hline
Baseline & - & $320.90$ \\
VideoFlow \cite{kumar} & $1.87$ & -   \\
SVP-FP \cite{denton2018stochastic}                 & - & $315.5$   \\
CDNA \cite{finn2016unsupervised}                   & - & $296.5$   \\
LVT (ours, $n_c=1$) & $1.25$ & $275.71\pm5.41$ \\ 
SV2P \cite{sv2p}                   & - & $262.5$   \\
LVT (ours, $n_c=4$) & $1.53$ & $125.8\pm2.9$ \\
SAVP \cite{savp}                    & - & $116.4$   \\
DVD-GAN-FP \cite{dvdgan}             & - & $109.8$   \\
TriVD-GAN-FP  \cite{dvdgan2}           & - & $103.3$   \\
Axial Transformer \cite{axial_transformer}     & 1.29 & -   \\
Video Transformer \cite{vt}       & 1.35 & $\mathbf{94\pm2}$ \\ \hline
\end{tabular}
\label{table:BAIRbench}
\end{table}

\begin{figure}[!hbtp]
\centering
\includegraphics[width=0.7\textwidth]{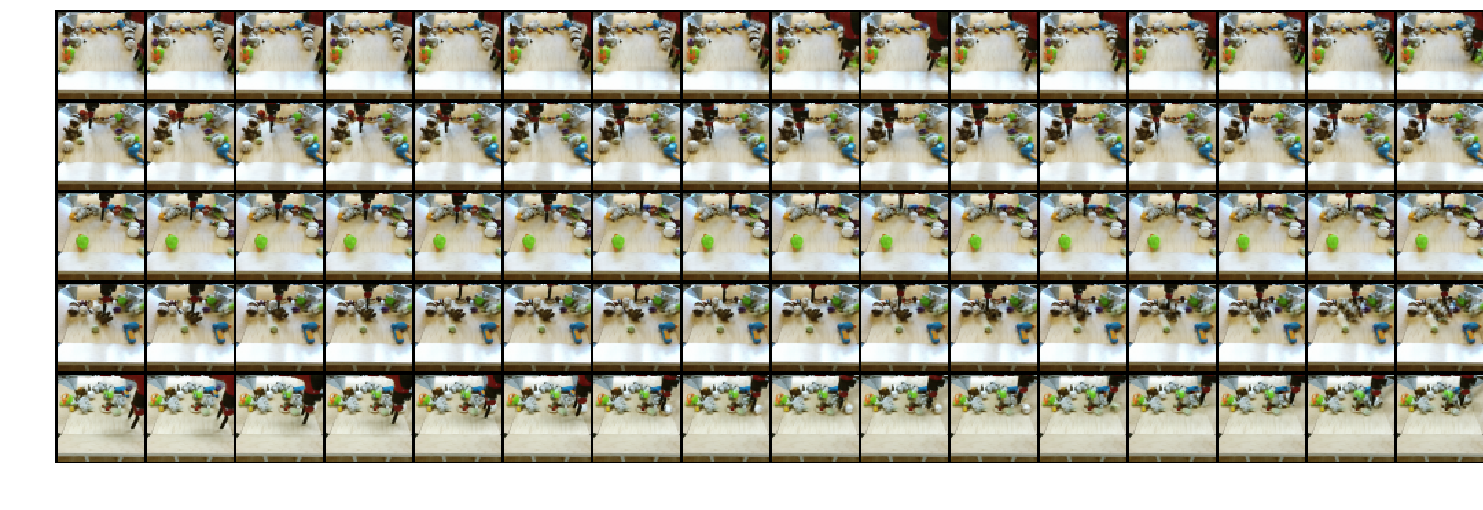}
\caption{Samples from BAIR Robot Pushing dataset. Each row represents a single video with first 5 frames being real and others generated.}
\label{fig:BAIR_samples}
\end{figure}

We achieve comparable performance in comparison to other methods. We also provide samples for qualitative assessment (see Fig \ref{fig:BAIR_samples}).

\subsection{Kinetics-600}


Kinetics-600 \cite{kinetics600_dataset} dataset consists of ~350k train and ~50k test videos. There are $600$ classes presented. Following the setup of previous approaches \cite{vt, dvdgan, dvdgan2}, we cropped each video to the size of the smallest side. Then we resized them to $64 \times 64$ using Lanczos filter. First, we evaluate the VQ-VAE's reconstruction error with varying number of codebooks used inside the discretization bottleneck. For that, we provide mean squared error (MSE) and FVD for reconstructed videos of 16 frames (see Table \ref{table:kinvqvae}).

\begin{table}[!hbtp]
\centering
\caption{VQ-VAE performance on Kinetics-600 dataset.}
\begin{tabular}{lccc}
\hline
$\mathbf{n_c}$ & \textbf{MSE}($\downarrow$) & \textbf{FVD($\downarrow$)}     \\ \hline
$1$ & $0.002$ & $396.58$ \\
$4$ & $0.0004$ & $25.95$ \\\hline
\end{tabular}
\label{table:kinvqvae}
\end{table}

VQ-VAE with four codebooks outperforms in terms of both metrics, and therefore later, we conduct the experiments under only this setup as the evaluation time for Kinetics-600 is particularly significant due to the large size of test data.

In terms of video prediction, following the setup of previous approaches \cite{vt, dvdgan, dvdgan2}, we train the video generator conditioning on one frame and report metrics for videos of 16 frames. FVD and bits/dim are computed on videos with five priming frames and one priming frame accordingly (see Table \ref{table:kinvt}).

\begin{table}[!hbtp]
\centering
\caption{Video prediction performance on Kinetics-600 dataset. Best results in bold.}
\begin{tabular}{lllccc}
\hline
\textbf{Subscaling type} & $\mathbf{n_c}$ & \textbf{Bits/dim}($\downarrow$) & \textbf{FVD}($\downarrow$) \\ \hline
Single Frame & 4 & $\textbf{2.14}$ & $\textbf{224.73}$ \\
Spatial & 4 &  $4.22$ & $2845.06 \pm 612.07$             \\
Spatiotemporal & 4 & $2.47$ & $338.39 \pm 0.21$ \\ \hline
\end{tabular}
\label{table:kinvt}
\end{table}

We compare our approach to others (see Table \ref{table:kinbench}). Here the baseline is the prediction of the next frame by the previous known frame.


\begin{table}[!hbtp]
\centering
\caption{Comparison of different methods for video prediction on Kinetics-600 dataset.}
\begin{tabular}{lcc}
\hline
\textbf{Method} & \textbf{Bits/dim}($\downarrow$)  & \textbf{FVD}($\downarrow$)     \\ \hline
Baseline & - & $271.00$ \\
LVT (ours) & $2.14$ & $224.73$ \\
Video Transformer \cite{vt} & $1.19$ & $170\pm5$ \\
DVD-GAN-FP \cite{dvdgan} & - & $69.15\pm1.16$   \\
TriVD-GAN-FP \cite{dvdgan2} & - & $25.74\pm0.66$ \\ \hline
\end{tabular}
\label{table:kinbench}
\end{table}

Our results are inferior to others on this dataset. We conclude that it is caused by error accumulation inside the Transformer model. We link it to the high complexity and diversity of the Kinetics-600 dataset. We want to emphasize that only four other approaches tried to model videos from this dataset, and all of them use six times bigger generative models (up to 350M parameters) than ours. In the meantime, increasing the size of our model led to a very slow convergence.

\begin{figure}[!hbtp]
\centering
\includegraphics[width=0.7\textwidth]{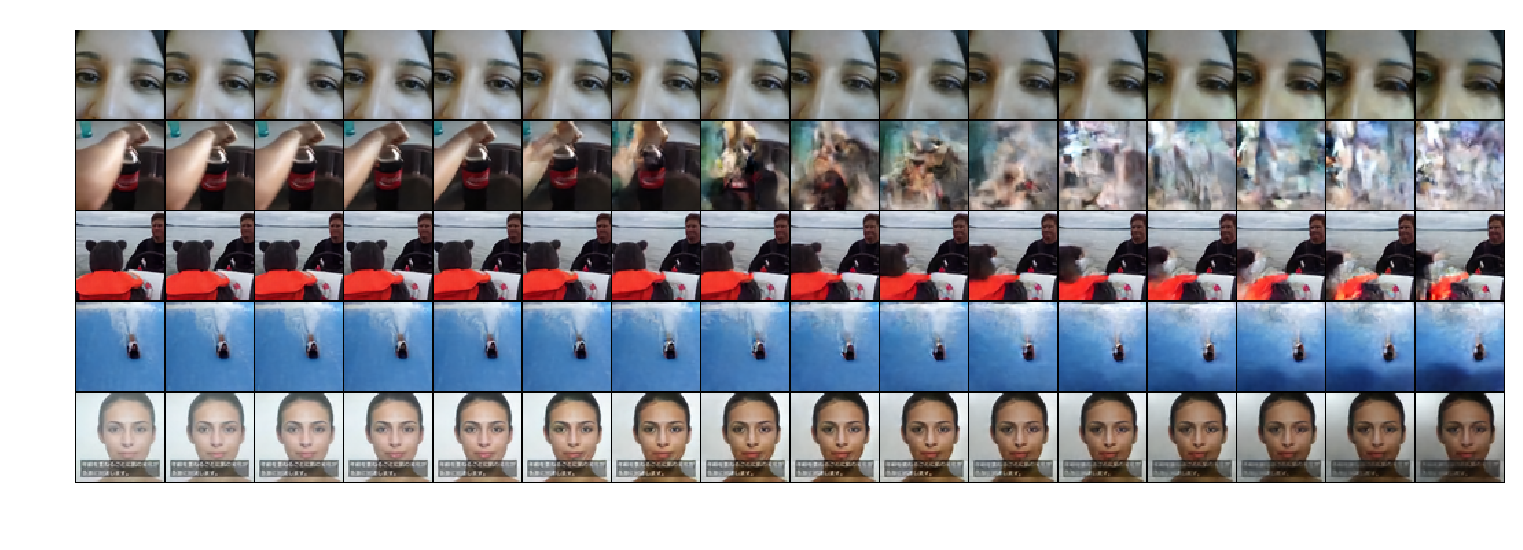}
\caption{Samples from Kinetics-600 dataset. Each row represents a single video with first five frames being real and others generated.} 
\label{fig:kin_samples}
\end{figure}

We also provide samples from our model for qualitative assessment (see Fig. \ref{fig:kin_samples}). One can notice artifacts in the second video. We found approximately half of the videos to be good and half of the videos to have artifacts. We provide more visualization results in the Appendix.

\subsection{Ablation Study of Attention}

The Video Transformer contains memory and time costly operation of multi-head attention. In general case multi-head attention computes feature $y_q$ as:

$$y_{q}=\sum_{m=1}^{M} W_{m}\left[\sum_{k \in \Omega_{q}} A_{m}\left(x_{q, }x_{k}\right) \odot V_{m} x_{k}\right],$$

where $m, q, k$ are the indexes of attention head, query and key elements respectively. $W_m$ and $V_m$ are learnable matrices. $A_m$ computes the attention weight for an each key element. We also normalize attention weights, s.t. $\sum_{k \in \Omega_{q}} A_{m}\left(q, k, z_{q}, x_{k}\right)=1$.

In our default setup, the attention weights are computed as:

$$A_{m}\left(x_{q}, x_{k}\right) \propto \exp \left(x_{q}^{\top} Q_{m}^{\top} K_{m} x_{k}+b_{kq}\right),$$

where $Q_{m}, K_{m}$ are learnable matrices for retrieving key and content embeddings, and $b_{kq}$ is computed as the sum of per-dimension relative distance biases between pixels $k$ and $q$. We would refer to this type of attention as \textit{``query-key + relative distance''}.

We also explore two other variants:
\begin{itemizetight}
    \item \textit{``key + relative distance'':} $A_{m}\left(x_{q}, x_{k}\right) \propto \exp \left(u_{m}^{\top} K_{m} x_{k}+b_{kq}\right)$, where $u_m$ is a lernable vector,
    \item \textit{``relative distance only'':} $A_{m} \left(x_{q}, x_{k}\right) \propto \exp \left(b_{kq}\right)$.
\end{itemizetight}

The empirical comparison of these different types of attention can be seen at Table \ref{table:attention}.

\begin{table}[!hbtp]
\centering
\caption{Attention comparison on BAIR Robot Pushing dataset. The results are obtained using the model with a single frame subscaling type modeling the latent space with $n_c=4$ codebooks.}
\begin{tabular}{lllccccc}
\hline
\textbf{attention type} & \textbf{Bits/dim}($\downarrow$) & \textbf{FVD}($\downarrow$) & \textbf{params (M)} & \textbf{inference time (sec)} \\ \hline
query-key + relative distance                    & $1.53$              & $125.76 \pm 2.90$ & $49.87$ & $35$  \\
key-only + relative distance                   & $1.57$              & $130.27\pm4.26$  & $41.50$ & $32$\\
relative distance only                    & $1.58$              & $141.62\pm4.34$  & $33.09$ & $30$\\ \hline
\end{tabular}
\label{table:attention}
\end{table}

Similar to \cite{zhu2019empirical} we find that query-key term inside self-attention module does not play a crucial role in the success of Latent Video Generator and can be replaced with cheaper variants with a cost of slight quality reduction.

\section{Conclusion}

In this work, we tackled the video generation problem. Given several first frames, the goal was to predict the continuation of a video. Modern methods for video generation requires up to 512 Tensor Processing Units for parallel training. We were focused on the reduction of the computational requirements of the model. We showed that one could achieve comparable results on video prediction by training a model using the usual research setup --- 8 V100 GPUs. To achieve such a result, we moved the video generation process from pixel space to a latent space. We demonstrated decent results on the dataset BAIR Robot Pushing. In the meantime, in some cases, we observe visual artifacts on the Kinetics-600 dataset.
\section*{Broader Impact}

In our work, we present a new model for video generation. Potentially, our model could be applied in cases where we need to predict the future. For example, prediction of further pedestrian movement on the photo, made by a camera on a self-driving car. Or animating photos of landscapes. One can use our model for animating pictures in their mobile phones, creating GIF animations.

We encourage researchers to understand and mitigate the risks of wrong future predictions. When the model is used in a self-driving car for pedestrian movement prediction, the error cost is high. Also, in some cases, there are long-tail events. If there is no such training data, the model could fail then. We suggest additional research to be done in terms of the reliability of our model.
\begin{ack}
The authors acknowledge the usage of the Skoltech CDISE HPC cluster Zhores for obtaining the results presented in this paper. The authors were supported by the Russian Science Foundation under Grant 19-41-04109. They also acknowledge Vage Egiazarian for thoughtful discussions of the model and the experiments.
\end{ack}

\medskip

\small

\bibliographystyle{plain}
\bibliography{papers.bib}

\newpage
\appendix

\section*{\uppercase{Appendix}}

\section{Adaptive Input and Adaptive Softmax.}

\begin{figure}[h!]
\centering
\includegraphics[width=0.9\textwidth]{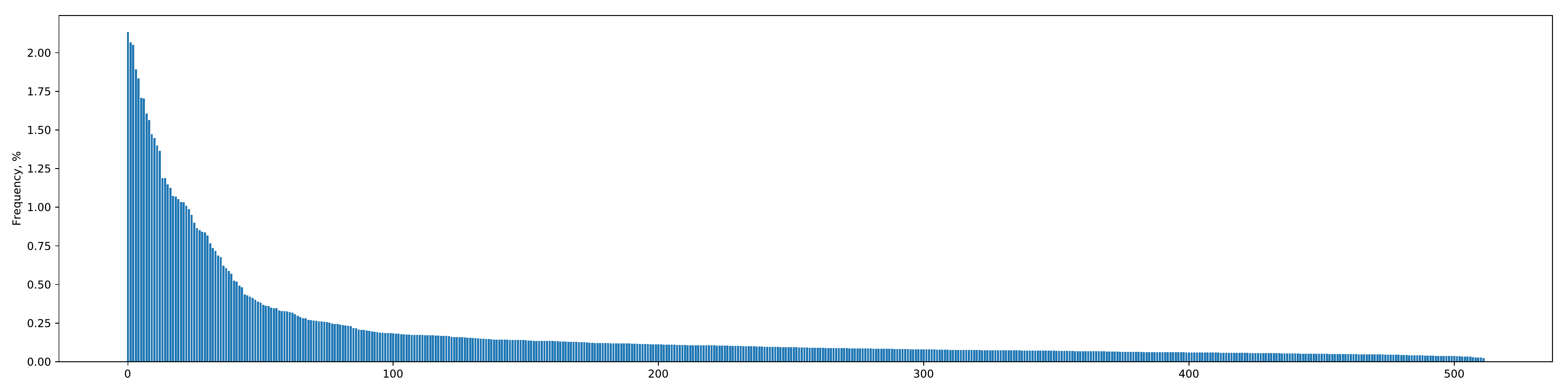}
\caption{Sorted codes frequencies for BAIR Robot Pushing dataset.} 
\label{fig:codebook_freq}
\end{figure}

We analyzed how often each latent code from the codebook was used for encoding images from train videos in BAIR Robot Pushing dataset. We found that 218 latent codes out of 512 constitute the 80\% of probability mass (see Fig. \ref{fig:codebook_freq}). Based on this fact, we tried to improve metrics using Adaptive Input \cite{adaptive_input} and Adaptive Softmax \cite{adaptive_softmax}. Neither of them brings an improvement to quality (see Table \ref{table:adaptive}), despite their successful applications in natural language processing. For an interested reader, we refer to the original papers for particular details.

\begin{table}[ht!]
\centering
\caption{Effects of applying adaptive input / softmax in decoder architecture. The results are obtained using the model with a single frame subscaling. Latent space is modelled with $n_c=1.$}
\resizebox{\textwidth}{!}{
\begin{tabular}{lllcccc}
\hline
\textbf{Subscaling type} & \textbf{Decoder Input} & \textbf{Decoder Output} & \textbf{Bits/dim}($\downarrow$) & \textbf{FVD}($\downarrow$) \\ \hline
Single Frame             & 128d Embedding         & 512d Softmax & $1.28$ & $\mathbf{258.89\pm2.85}$  \\
Spatial                  & 128d Embedding         & 512d Softmax & $1.79$ & $524.43\pm9.41$             \\
Spatiotemporal            & 128d Embedding         & 512d Softmax & $\mathbf{1.25}$ & $275.71\pm5.41$  \\ \hline
Single Frame             & 128d Embedding         & 512d Softmax (tied emb) & $1.27$ & $265.10\pm3.85$  \\
Single Frame             & Adaptive Input         & 512d Softmax & $1.26$ & $259.73\pm6.35$              \\
Single Frame             & 128d Embedding         & Adaptive Softmax & $1.37$ & $265.99\pm4.16$              \\
Single Frame             & Adaptive Input         & Adaptive Softmax & $1.36$ & $259.88\pm5.25$              \\
Single Frame             & Adaptive Input         & Adaptive Softmax (tied emb) & $1.35$ & $259.55\pm8.50$              \\
Single Frame             & Adaptive Input         & Adaptive Softmax (tied emb/proj) & $1.34$ & $264.79\pm4.27$              \\ \hline
\end{tabular}}
\label{table:adaptive}
\end{table}

\section{BAIR visualizations}

\begin{figure}[!h]
\centering
\includegraphics[width=0.9\textwidth]{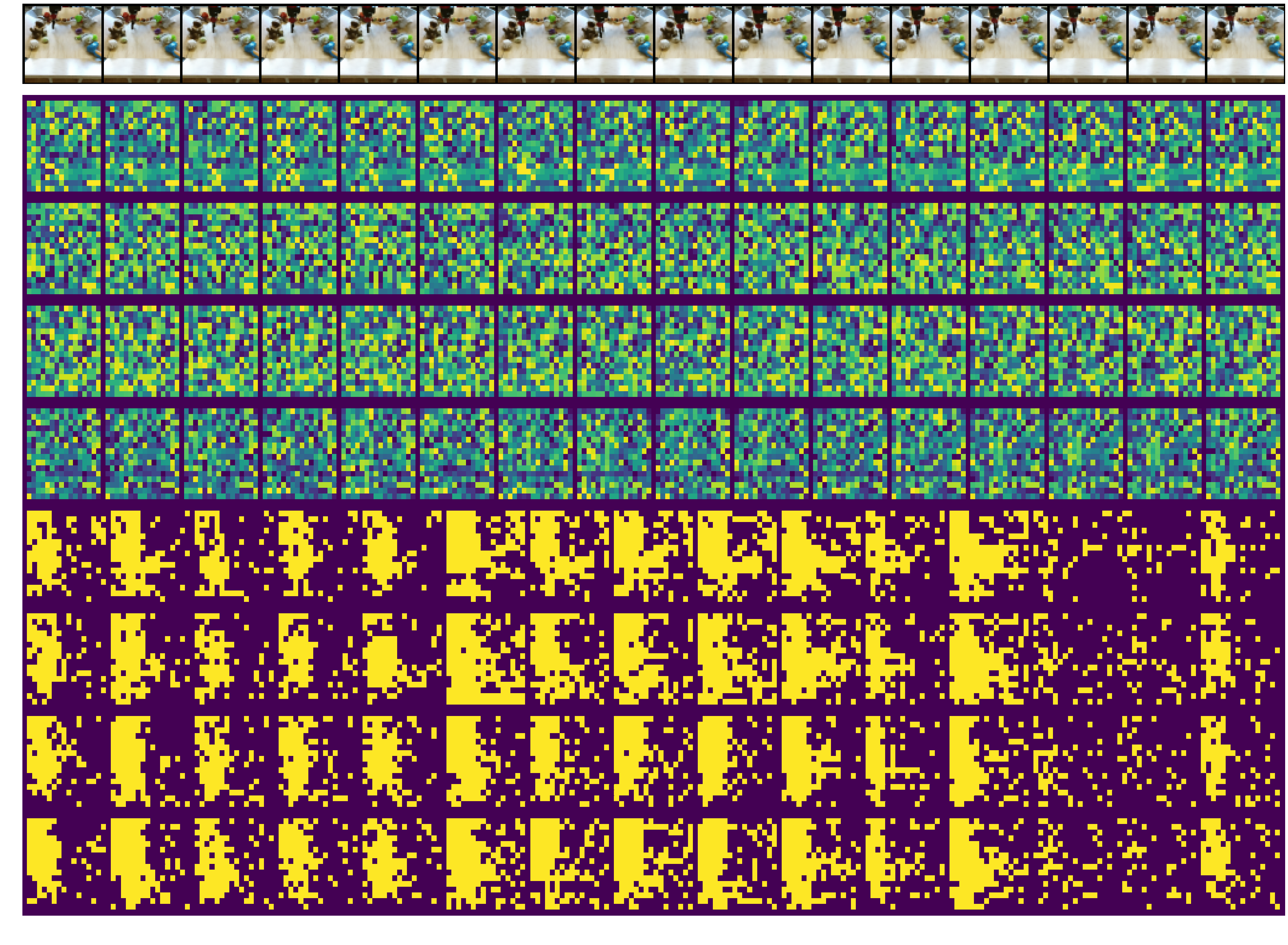}
\caption{Sample from BAIR robot pushing dataset. The first row represents a single video with the first five frames being real and others generated. Rows 2-5 represent four latent codes, one row for each codebook. Rows 6-9 represent binary mask denoting whether the latent code between consecutive frames changes or not (yellow means a change).}
\label{fig:bair_codes}
\end{figure}

\begin{figure}[!h]
\centering
\includegraphics[width=0.9\textwidth]{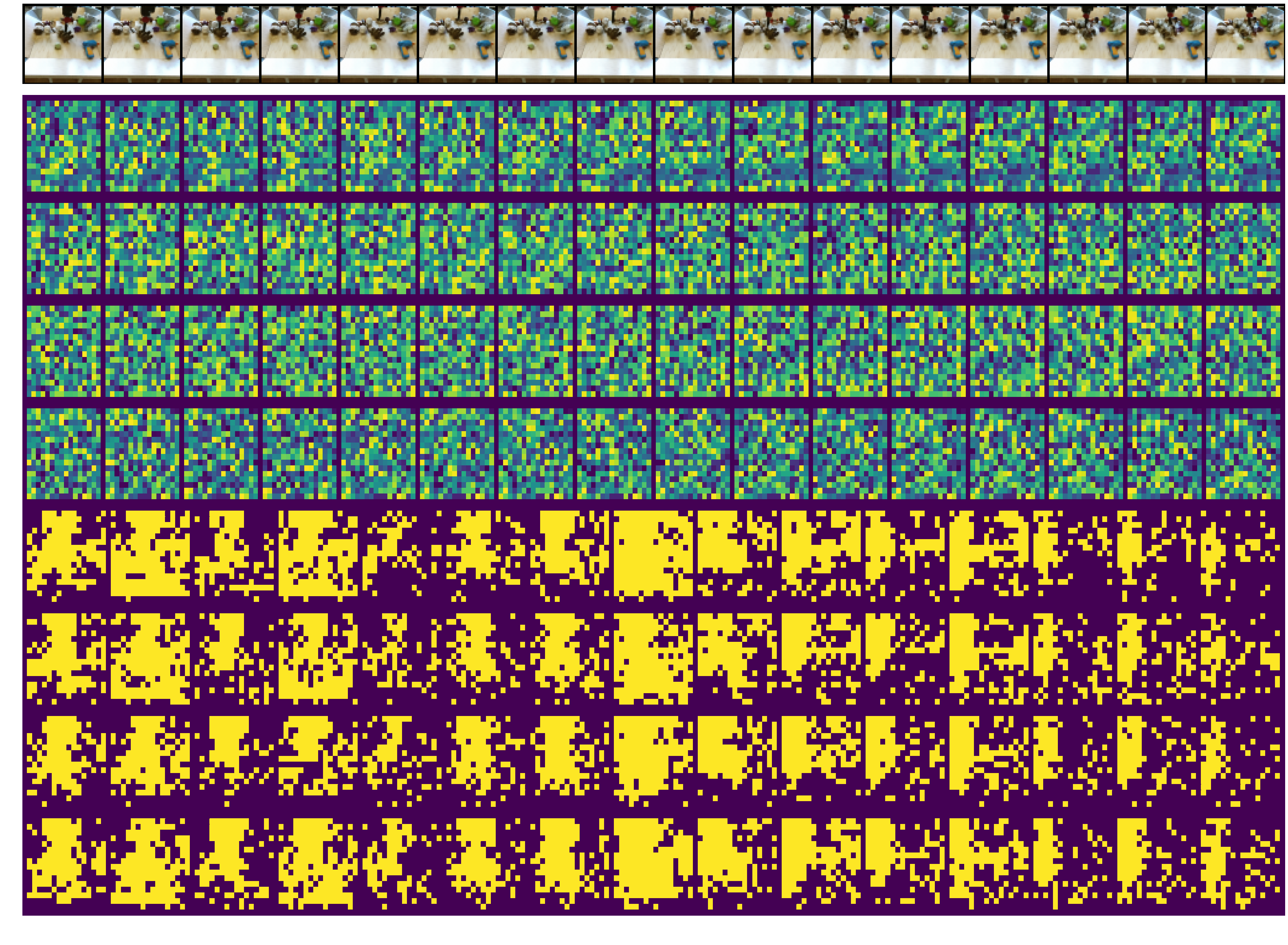}
\caption{Sample from BAIR robot pushing dataset. The first row represents a single video with the first five frames being real and others generated. Rows 2-5 represent four latent codes, one row for each codebook. Rows 6-9 represent binary mask denoting whether the latent code between consecutive frames changes or not (yellow means a change).} 
\label{fig:bair_codes2}
\end{figure}

We present visualizations of BAIR robot pushing dataset in Fig. \ref{fig:bair_codes} and Fig. \ref{fig:bair_codes2}. At the first row of each figure there is a video: $5$ real and $11$ generated frames. Rows 2-5 contain codes visualizations. Each code row corresponds to one codebook. Since a single code is just a matrix of indexes, we decode it with a technique called \textit{indexed color}. In other words, we assigned each index in a code to a specific color. Rows 6-9 represent binary mask denoting whether the latent code between consecutive frames changes or not (yellow means a change).
 
One can see that results on the BAIR dataset are quite realistic. Also, it is important to note that some of the codes stay the same from frame to frame. The static background causes it. Codes that are responsible for non-static objects change from frame to frame.

\section{Kinetics-600 visualizations}

\begin{figure}[!h]
\centering
\includegraphics[width=\textwidth]{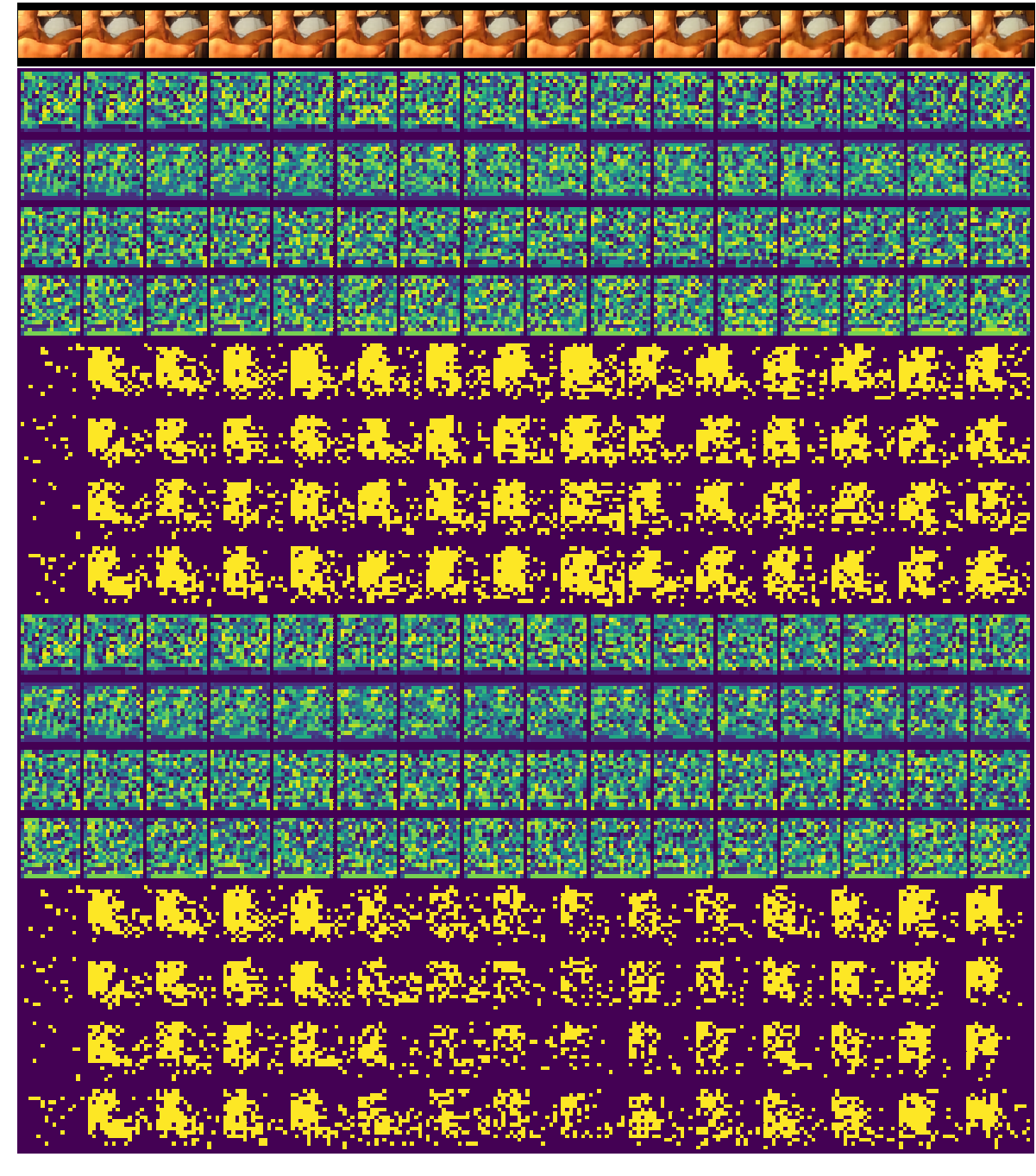}
\caption{Good sample from Kinetics-600 dataset. The first row represents a single video with the first five frames being real and others generated. Rows 2-5 represent four latent codes for real video, one row for each codebook. Rows 6-9 represent binary mask denoting whether the latent code between consecutive frames changes or not (yellow means a change). Rows 10-13 represent four latent codes for generated video, one row for each codebook. Rows 14-17 represent binary mask denoting whether the latent code between consecutive frames changes or not (yellow means a change).}
\label{fig:kin_good_1}
\end{figure}

\begin{figure}[!h]
\centering
\includegraphics[width=\textwidth]{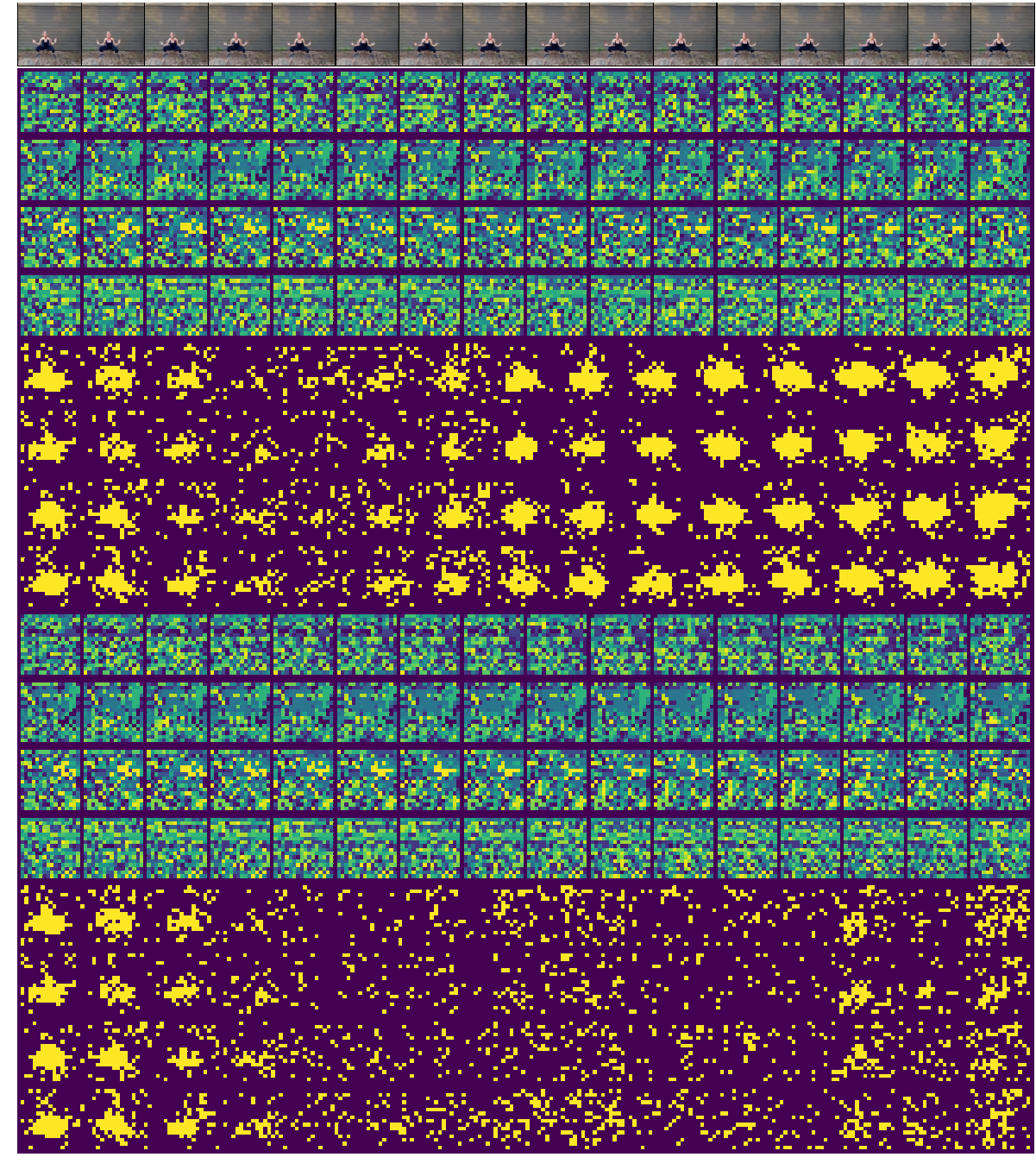}
\caption{Good sample from Kinetics-600 dataset. The first row represents a single video with the first five frames being real and others generated. Rows 2-5 represent four latent codes for real video, one row for each codebook. Rows 6-9 represent binary mask denoting whether the latent code between consecutive frames changes or not (yellow means  a change). Rows 10-13 represent four latent codes for generated video, one row for each codebook. Rows 14-17 represent binary mask denoting whether the latent code between consecutive frames changes or not (yellow means a change).} 
\label{fig:kin_good_2}
\end{figure}

\begin{figure}[!h]
\centering
\includegraphics[width=\textwidth]{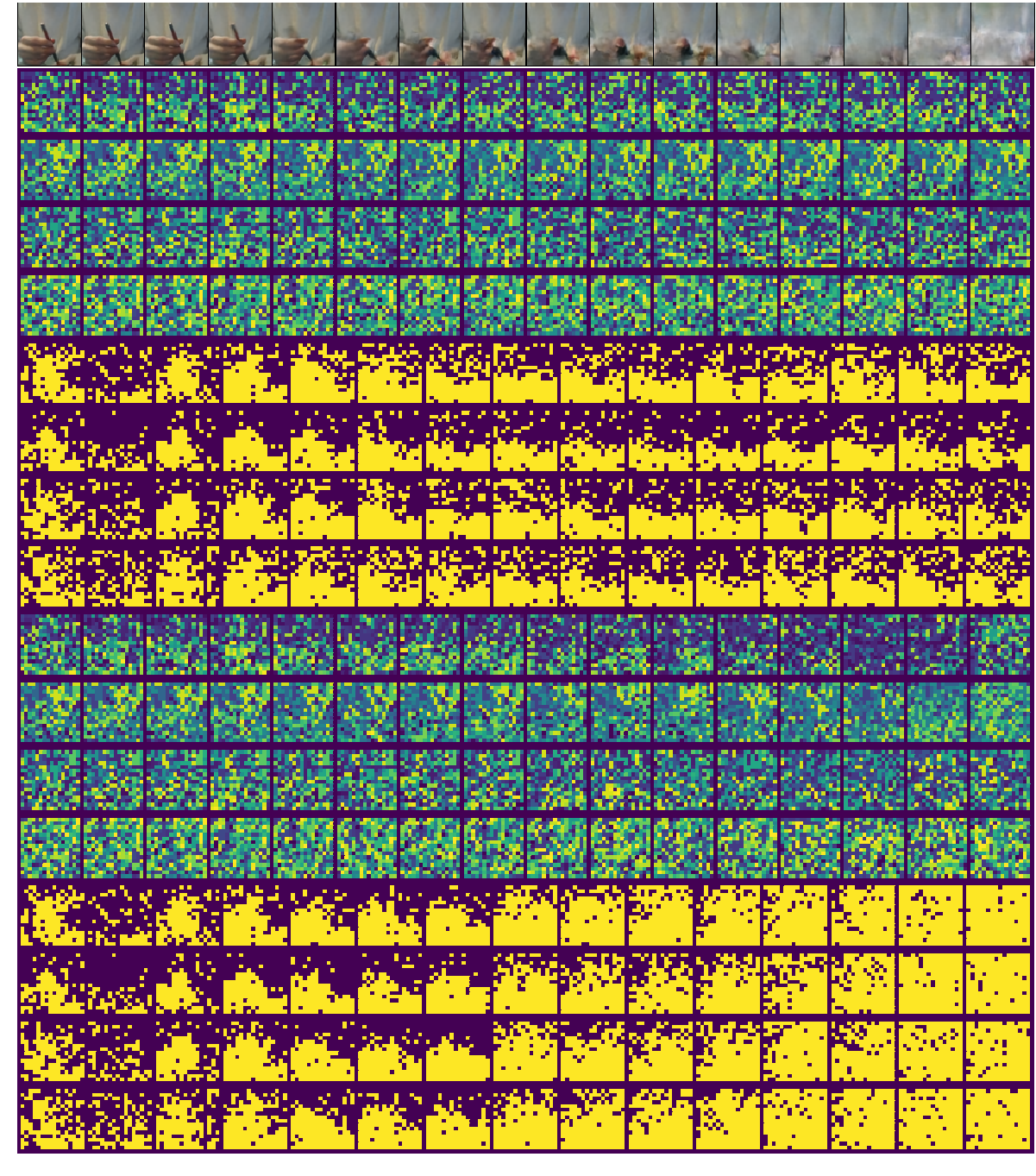}
\caption{Bad sample from Kinetics-600 dataset. The first row represents a single video with the first five frames being real and others generated. Rows 2-5 represent four latent codes for real video, one row for each codebook. Rows 6-9 represent binary mask denoting whether the latent code between consecutive frames changes or not (yellow means a change). Rows 10-13 represent four latent codes for generated video, one row for each codebook. Rows 14-17 represent binary mask denoting whether the latent code between consecutive frames changes or not (yellow means a change). }
\label{fig:kin_bad_1}
\end{figure}

\begin{figure}[!h]
\centering
\includegraphics[width=\textwidth]{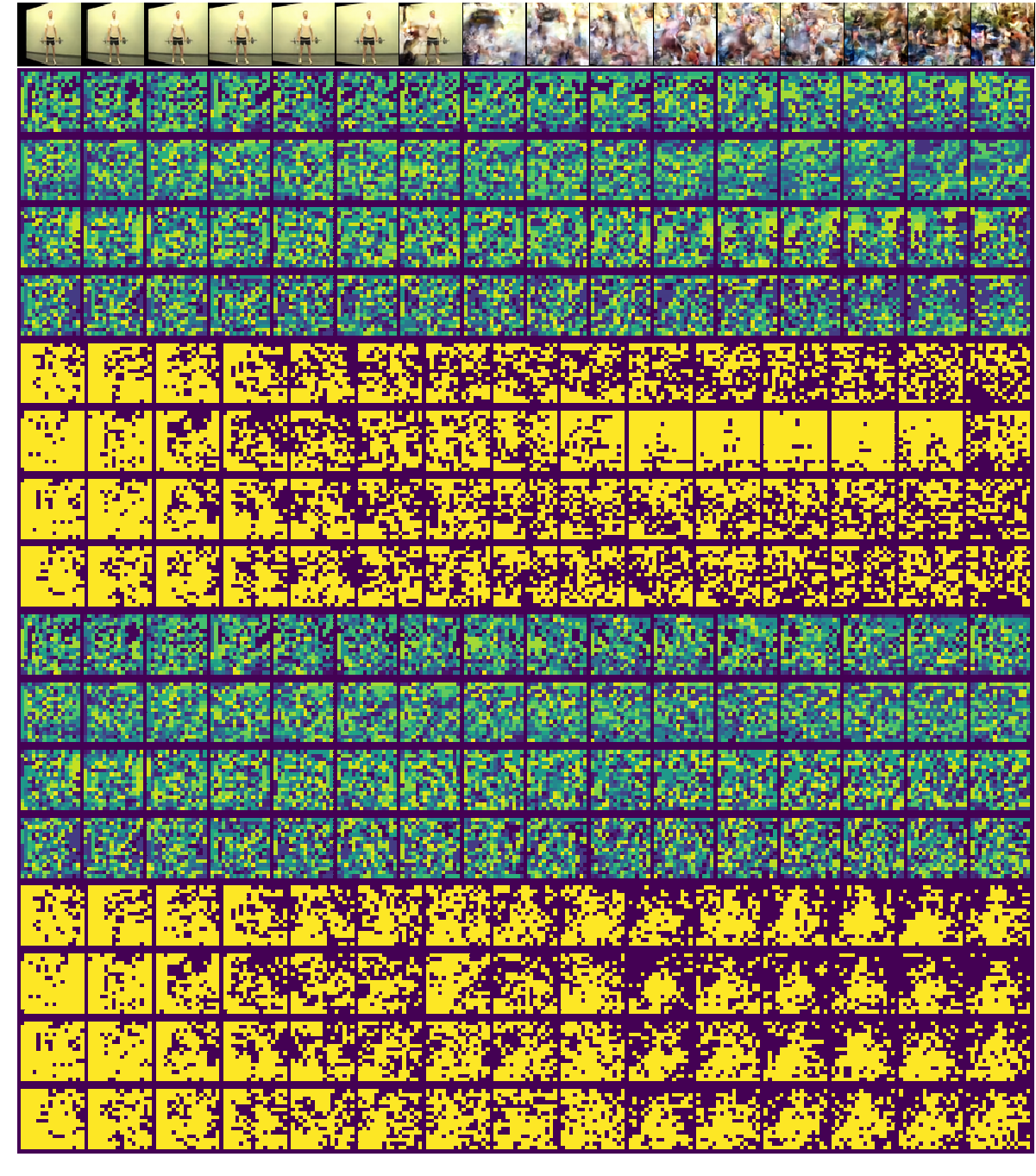}
\caption{Bad sample from Kinetics-600 dataset. The first row represents a single video with the first five frames being real and others generated. Rows 2-5 represent four latent codes for real video, one row for each codebook. Rows 6-9 represent binary mask denoting whether the latent code between consecutive frames changed or not (yellow means changed). Rows 10-13 represent four latent codes for generated video, one row for each codebook. Rows 14-17 represent binary mask denoting whether the latent code between consecutive frames changes or not (yellow means a change).} 
\label{fig:kin_bad_2}
\end{figure}

We present good samples along with codes on the Kinetics-600 dataset in Figures \ref{fig:kin_good_1} and \ref{fig:kin_good_2}. Bad samples with codes are presented in Fig. \ref{fig:kin_bad_1} and \ref{fig:kin_bad_2}. One can see that bad samples and good samples are different in their codes and code differences. For latent transformer, it is easier to predict the latent code of the next frame if it is similar to the latent code of the current frame.

\begin{figure}[!h]
\centering
\includegraphics[width=\textwidth]{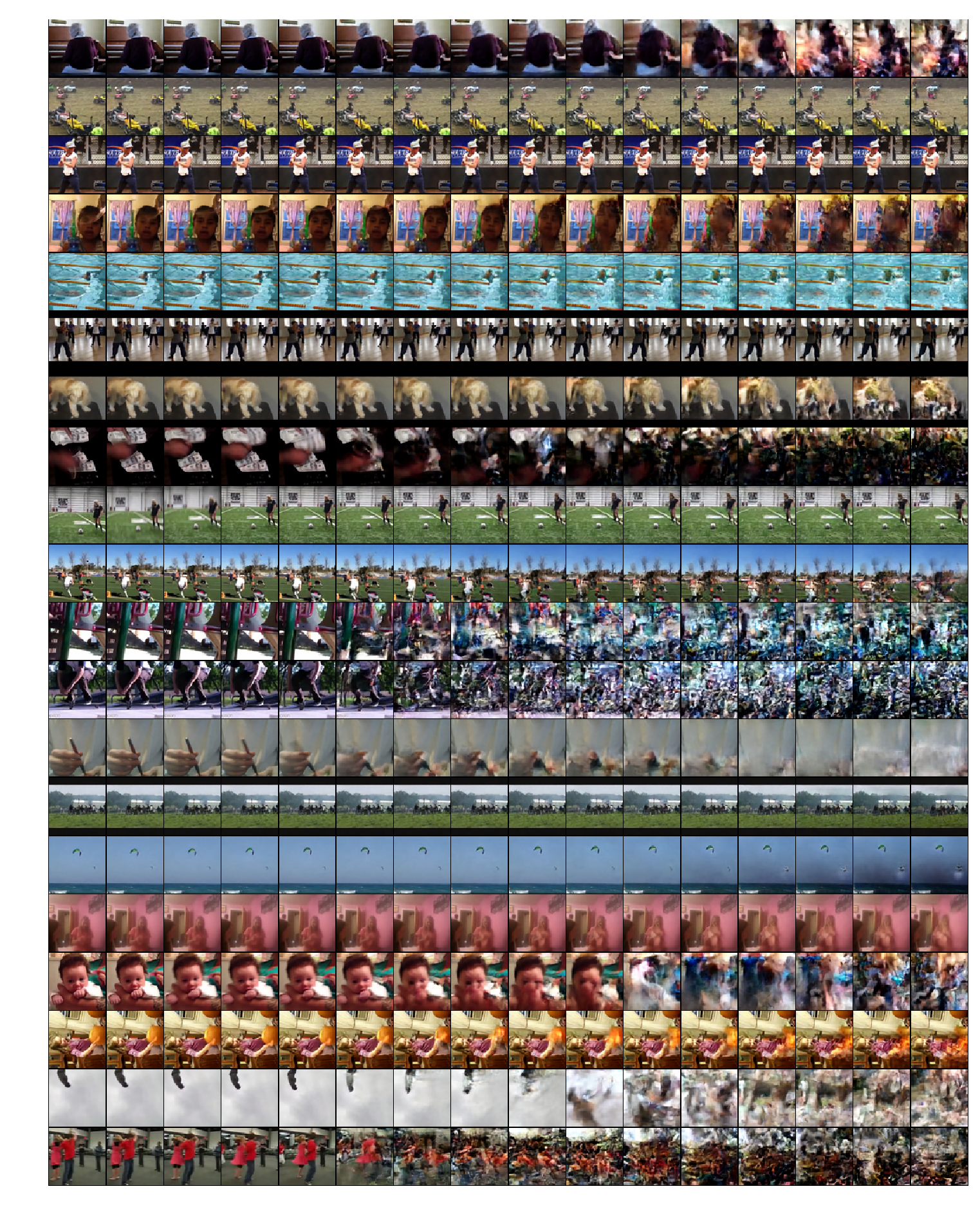}
\caption{Samples from Kinetics-600 dataset. Each row represents a single video with first five frames being real and others generated.} 
\label{fig:kinetics_many_20}
\end{figure}

More generated samples are presented in Fig. \ref{fig:kinetics_many_20}. For each video, first five frames are real, others are generated.

\end{document}